# Efficient Multi-Slide Visual-Language Feature Fusion for Placental Disease Classification

Hang Guo*
East China Normal University
Shanghai, China
51285904046@stu.ecnu.edu.cn

Qing Zhang*
East China Normal University
Shanghai, China
qzhang@ce.ecnu.edu.cn

Zixuan Gao
East China Normal University
Shanghai, China
10232140401@stu.ecnu.edu.cn

Siyuan Yang
Nanyang Technological University
Singapore, Singapore
siyuan005@ntu.edu.sg

Shulin Peng
East China Normal University
Shanghai, China
10225304467@stu.ecnu.edu.cn

Xiang Tao
Obstetrics and Gynecology Hospital
of Fudan University
Shanghai, China
taoxiang1696@fckyy.org.cn

Ting Yu
Obstetrics and Gynecology Hospital
of Fudan University
Shanghai, China
yuting7410@fckyy.org.cn

Yan Wang
East China Normal University
Shanghai, China
ywang@cee.ecnu.edu.cn

Qingli Li†
East China Normal University
Shanghai, China
qlli@cs.ecnu.edu.cn

## Abstract

Accurate prediction of placental diseases via whole slide images (WSIs) is critical for preventing severe maternal and fetal complications. However, WSI analysis presents significant computational challenges due to the massive data volume. Existing WSI classification methods encounter critical limitations: (1) inadequate patch selection strategies that either compromise performance or fail to sufficiently reduce computational demands, and (2) the loss of global histological context resulting from patch-level processing approaches. To address these challenges, we propose an **E**fficient **m**ulti**m**odal framework for **P**atient-level placental disease **D**iagnosis, named **EmmPD**. Our approach introduces a two-stage patch selection module that combines parameter-free and learnable compression strategies, optimally balancing computational efficiency with critical feature preservation. Additionally, we develop a hybrid multimodal fusion module that leverages adaptive graph learning to enhance pathological feature representation and incorporates textual medical reports to enrich global contextual understanding. Extensive experiments conducted on both a self-constructed patient-level Placental dataset and two public datasets demonstrating that our method achieves state-of-the-art diagnostic performance. The code is available at https://github.com/ECNU-MultiDimLab/EmmPD.

*Both authors contributed equally to this research.
†Corresponding author.

## 1 Introduction

Accurate prediction of placental diseases is essential for reducing the risk of life-threatening complications in both mothers and newborns. Early and reliable diagnosis enables timely clinical interventions and better pregnancy management. While slide image (WSI) classification plays a vital role in this process, as it allows for comprehensive analysis of placental tissue morphology at the patient level. Digital pathology enables large-scale analysis of WSIs, providing a powerful foundation for accurate and scalable prediction of various diseases [20, 25, 41, 49]. Though existing methods excel with sufficient data, placental digital pathology remains limited due to data acquisition challenges. Current research mostly focuses on regional placental analysis. For example, Happy [47] develops a hierarchical framework for quantification assessment of pathological cells and micro-anatomical tissue structures in placental WSI. GestAltNet [31] proposes a attention based deep learning model to predict gestational age from placental WSI. Jeffery *et al.* applied machine learning to identify and classify three types of placental parenchymal lesions [11]. Although these pathology-based methods provide valuable auxiliary diagnostic evidence for pathologists, they cannot yet deliver conclusive diagnostic determinations for placental diseases. Current deep learning studies on placental disease focus on MRI or ultrasound [33, 54], while histopathology—the gold standard—remains underexplored. One patient may present with multiple placental diseases that are often inadequately represented in a single WSI, which may even appear as a negative sample. Additionally, the absence or insufficient number of chorionic villi in a single slide can significantly influence diagnostic accuracy [45]. Consequently, pathologists typically prepare multiple cross-sectional slides per patient and perform comprehensive analysis across these slides to reach a definitive placental disease diagnosis. WSIs contain billions of pixels, imposing heavy computational demands. Combining multiple WSIs for analysis generates substantially greater data volume, dramatically increasing computational complexity. Thus, Multiple Instance Learning (MIL) is widely used in current approaches and effective patch selection remains a key research focus. As shown in Fig. 1, current patch selection strategies primarily follow two categories: traditional and deep learning methods. Traditional methods reduce WSI data by removing tissue-free regions [15] or applying dimensionality reduction techniques [17]. Deep learning methods typically extract patch-level image features through pre-trained models, followed by the selection of representative patches through clustering [50] or attention mechanisms [56]. Although these methods can partially reduce computational requirements, they remain constrained by two critical limitations: inadequate compression ratios or excessive compression that adversely impacts model performance.

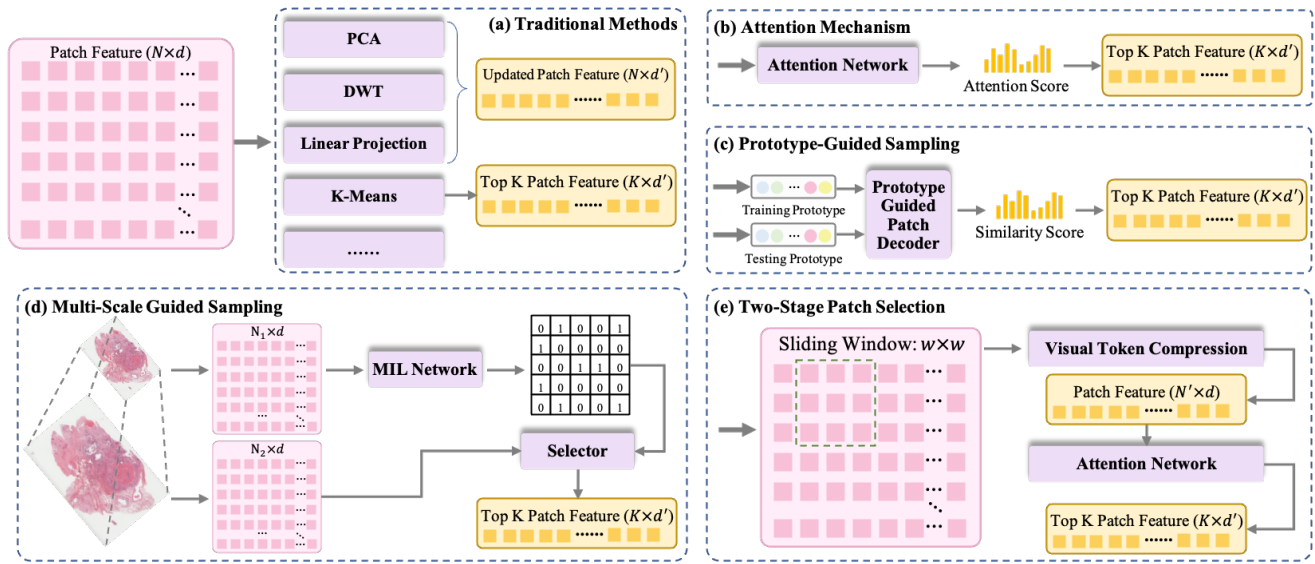

Figure 1: Comparison of different patch selection strategies from WSI data.

While WSI data compression technology effectively reduces computational demands, its predominantly patch-level implementation inevitably results in the loss of global histological features. To compensate for these visual feature deficiencies, pathological reports have emerged as a valuable supplementary modality. This has led to significant development of vision-language models for WSI classification, which generally fall into three categories: 1) simple concatenating, 2) cross-modal attention, and 2) knowledge-distillation. Simple concatenation approaches merge visual and textual features at either low-level or middle-level stages [20]. While computationally efficient, these methods overlook the deep correlations between the two modalities. Cross-modal attention-based methods employ transformer-based architectures with co-attention mechanisms to enable interaction between visual and textual features [29]. While these models can capture dynamic and context-aware associations, their computational cost escalates significantly in multi-slide settings, and they typically require large-scale annotated datasets. Knowledge-distillation approaches utilize pathology reports to guide the learning of visual feature [3]. These approaches effectively filter redundant image regions using text supervision but rely heavily on its quality.

To address the aforementioned challenges, we propose a computationally **E**fficient **m**ulti**m**odal framework for multi-slide joint analysis in **P**atient-level placental disease **D**iagnosis, named EmmPD. To overcome the WSI data volume challenge, we develop a novel two-stage hybrid compression module that synergistically combines parameter-free compression with a trainable compression component. This architecture effectively selects key diagnostic regions, achieving an optimal balance between compression and computational. Given that patch-level processing strategy risks omitting clinically critical macroscopic patterns, we further introduce a hierarchical multimodal fusion module to extract multi-grained pathological features. In the first stage, we introduce an adaptive graph learning mechanism that further enhances information representation through graph-based learning of compressed patches. Subsequently, considering the regionally distributed nature of pathological data, we incorporate textual descriptions to guide the visual-language feature fusion, thereby enriching global contextual understanding. We collect pathological reports and multiple slides from patients with placental diseases, and construct a visual-language dataset for experimental research. Experimental results demonstrate that our proposed EmmPD achieves state-of-the-art diagnostic accuracy for placental disease while maintaining computational efficiency. Our contributions are as follows:

- We propose an efficient multi-slide analysis framework to address computational challenges in WSI classification.
- We introduce a two-stage patch compression architecture to extract the most important patches strongly associated with different placental diseases.
- We design a hybrid multi-grained multimodal feature fusion module that incorporates graph-based learning for enhanced pathological representation, and employing textual descriptions to compensate for global features.
- We establish a multi-slide placental disease classification dataset containing WSIs and pathological reports. Experiments conducted on the home-made Placenta dataset and two public datasets demonstrate the superiority of EmmPD.

## 2 Related Work
### 2.1 Multiple Instance Learning in WSI

Given the gigapixel scale of WSIs, MIL has advanced the field by capturing both local and global features, enhancing performance in object localization [9, 39] and classification [6, 41]. In MIL-based models, each slide is treated as a bag of patches with a slide-level label, enabling learning without pixel- or patch-level annotations. These models can be classified into four groups: transformer-based, graph-based, self-supervised, and multimodal methods. Transformer-based methods like TransMIL [36] and DTFD-MIL [55] use transformer architectures to capture complex intra-slide relationships. Graph-based models represent WSIs as graphs with patches as nodes and semantic relations as edges. For instance, SlideGCD [43] builds slide-level graphs to mine inter-slide correlations. Self-supervised methods like HIPT [4] and Virchow [48], combine feature pretraining with MIL for downstream tasks, though they require large datasets. Multimodal MIL approaches fuse information from different modalities—e.g., image and text [57] or multi-resolution inputs [27]—to enrich representation. While most methods focus on patch-level feature aggregation for slide-level classification, few explore joint analysis across multiple slides. Among those that do, such as [13, 19, 43], efforts are usually limited to a single disease type. To the best of our knowledge, patient-level WSI analysis remains largely underexplored. HVTSurv [37] aggregates slide- and patient-level features for survival prediction. However, no existing work addresses patient-level disease diagnosis, where multiple slides from one patient may involve multiple disease types.

### 2.2 Pathological Patch Selection

WSI analysis methods require large computational resources due to its high resolution with gigapixel-level data. To alleviate this, both traditional and learnable data compression strategies have been proposed (see Fig. 1). Traditional methods rely on color, texture, and morphological analysis to discard background and redundant patches—for example, HistoQC [16] and PCA-DWT [17]. These methods are computationally efficient but offer limited compression. Learnable patch selection dynamically identifies informative patches during inference, enhancing efficiency. Attention-based methods, like CLAM [30], assign weights to patches to highlight key regions. Prototype-based methods, like PANTHER [44], PAM [12], and Pamil [26], select test patches aligned with training prototypes. Multi-scale methods, such as CoD-MIL [40] and DAS-MIL [2], enhance sampling by using coarse-scale guidance to improve fine-scale selection. While these methods reduce computational load, their overall compression ratios remain limited.

### 2.3 Multimodel Feature Fusion Methods

Multimodal feature fusion has gained traction in computational pathology, boosting diagnostic accuracy and enhancing model robustness to data variability. Recent approaches include feature-, cross-modal, and decision-level fusion. Feature-level fusion operates at the input or shallow layers by directly combining raw data. MMMI [21], which uses a nonlinear layer for early fusion but lacks flexibility and fails to capture multimodal synergy. Decision-level fusion methods like DDEF [38] and MSFFS [58] process each modality independently and fuse at the decision stage. While flexible and easy to implement, they miss deeper inter-modal interactions and incur high computational costs. CLIP [35] aligns image-text embeddings via contrastive learning for effective cross-modal fusion. LMAM [42] uses a lightweight module to assign matching weights and compute inter-modal attention scores. Cross-modal fusion strategy can capture inter-modal interactions and enhance model's representation ability [1, 8, 51]. CrossFormer [23] proposes an effective heterogeneous graph transformer module for cross-modal representation learning. Despite their effectiveness, they have high complexity and high computation cost.

## 3 Method
### 3.1 Pipeline Overview

WSI classification is typically formulated as a multiple instance learning problem due to the extremely high resolution of the slides. In this paper, we collect a patient-level pathological dataset, where each patient has multiple WSIs $X$ and a single associated pathology report. We define the input as image-text pairs $\mathcal{D} = (X, T)$, where $T$ is the corresponding disease description. The goal is to predict the patient-level label $\hat{Y} \in 1, 2, ..., C$, where $C$ is the number of disease classes. All WSIs of a patient are divided into a patch-wise bag $B$ with $N$ patches and a bag-level label $Y$ is derived from the pathology report $T$. Each patch is embedded using the image encoder of pre-trained pathology foundation model UNI [5], denoted as $\mathcal{E}_{UNI}$. This yields a set of $d$ dimensional patch-level feature embeddings:

$$B = \{\mathcal{E}_{UNI}(x_i) : x_i \in X\} = \{v_i\}_{i=1}^{N}, \qquad (1)$$

where $\{v_i \in \mathbf{R}^d\}$ is the extracted visual feature for the $i^{th}$ patch.

To reduce computation, we adopt a two-stage patch compression: redundancy removal via spatial-semantic similarity, followed by attention-based key patch selection. While patch-level features capture local details, they may miss global and inter-slide structural context. To address this, we further incorporate textual reports to enrich global semantics and construct a patch-level graph across WSIs using a GCN to capture spatial-structural dependencies. Finally, we propose a hybrid fusion strategy combining visual features, disease text, and GCN-enhanced structural information. This tri-modality fusion results in a discriminative bag-level embedding that captures both local morphological details and global contextual information essential for WSI classification.

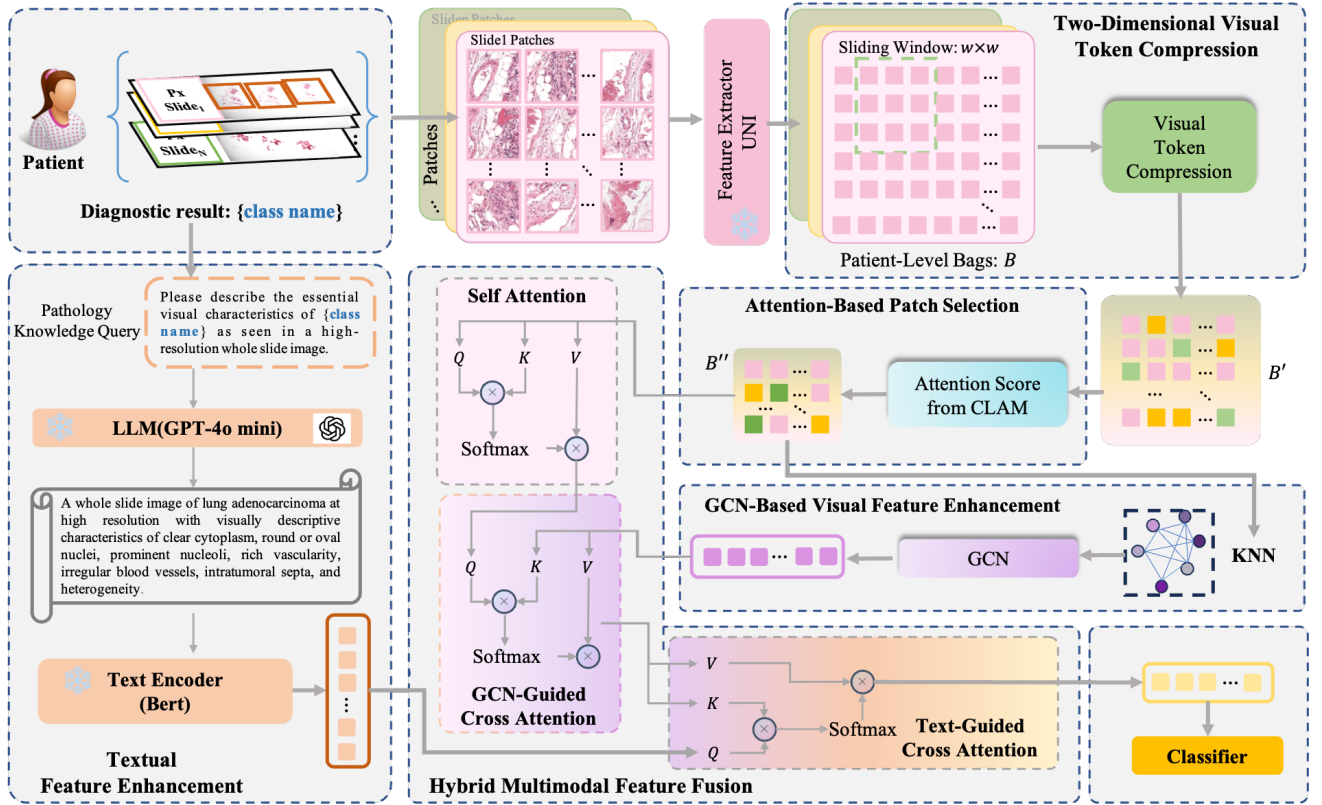

Figure 2: Illustration of our proposed architecture. The input WSIs are cropped into patches and fed into a feature extractor to get a patient-level feature bag $B$. Then $B$ is processed with a two stage patch selection module, including two-dimensional visual token compression and attention-based patch selection. At last, the updated feature bag $B''$ is progressively enhanced by graph and textual information to generate accurate patient-level disease classification results.

## 3.2 Two-Stage Patch Selection

The gigapixel scale of WSIs poses major computational challenges, and simple patch sampling or channel reduction is inadequate. Since adjacent patches are highly similar, we propose a two-stage patch selection to first eliminate redundancy, then select the most informative patches for classification.

*3.2.1 Two-Dimensional Visual Token Compression.* WSIs contain vast numbers of patches, causing high computational costs even after compression. Studies show only 10% are diagnostically relevant, while 90% are redundant and distract the model from key areas. To address this, we propose a two-dimensional visual token compression method as a preliminary step to eliminate redundant patches and reduce computational load. We compress the features extracted by UNI [5] and determine the $k$-nearest neighbors of each patch based on their coordinate information, *i.e.*, selecting $k$ adjacent patches within a sliding window centered on a given patch. For each two-dimensional window of size $w \times w$, we obtain the $i^{th}$ normalized feature $\hat{v}_i = \frac{v_i}{||v_i||_2}$, and then compute pairwise similarity between $\hat{v}_i$ and $\hat{v}_j$.

$$S_{ij} = \hat{v}_i \cdot \hat{v}_j, i \neq j. \quad (2)$$

To adapt to varying tissue structures across different slides, we define a dynamic threshold $\theta$ as the mean similarity within each window: $\theta = \frac{1}{w^2} \sum_{i=1}^{w^2} S_{ij}$. If the average similarity of a patch-pair exceeds $\theta$, the pair is considered redundant and then one of the patches is removed, generating a compressed patch bag $B'$. Utilizing a two-dimensional window, this module serves as a coarse-grained filter to efficiently eliminate redundant patches with highly repetitive information in the WSI, thereby reducing computational overhead and improving analysis efficiency. Serving as an initial screening mechanism before more advanced compression techniques, it helps preserve the informative regions while discarding unnecessary data.

*3.2.2 Attention-Based Patches Selection.* Although visual token compression removes redundancy, some retained patches may lack diagnostic value. To address this, we adopt an attention-based selection module (inspired by CLAM [30]) that scores patches by their diagnostic relevance. The top-$K$ patches with the highest attention scores are selected from the compressed bag $B'$ as the input for subsequent modules. Combing the aforementioned two compression blocks, we can efficiently compress the data to an updated patch-level bag $B'' = \{v_i\}_{i=1}^{K}$.

### 3.3 Hybrid Multimodal Feature Fusion

*3.3.1 Textual Feature Enhancement.* Traditional MIL-based WSI classification methods lack global context, limiting their ability to capture broader diagnostic patterns. To mitigate this, we incorporate textual descriptions as global medical knowledge to bridge the visual-semantic gap. Therefore, for each disease mentioned, we develop a pathological knowledge query: *"Please describe the essential visual characteristics of {class name} as seen in a high-resolution whole slide image"*. Next, we utilize ChatGPT-4o mini [32] generate detailed diagnostic criteria and morphological features. These texts are encoded using the frozen BERT text encoder [7], extracting the [CLS] token embedding as the textual feature representation $T_f \in \mathbb{R}^{C \times d}$. To enhance alignment with visual features, we concatenate $T_f$ with a learnable prompt embedding $T_l \in \mathbb{R}^{t \times d}$, forming the final textual prompt: $T = [T_f, T_l] \in \mathbf{R}^{(C+t) \times d}$. By incorporating learnable prompt tokens, EmmPD can be more flexible and generalized in WSI classification task.

*3.3.2 GCN-Based Visual Feature Enhancement.* Patch-level analysis may miss spatial structural relationships. To address this, we build a patch-level graph $G = (V, E)$ from the refined patch set $B''$, where $V$ denotes the set of patch features, and $E$ represents the set of edges that connect patches. Processed by our proposed two-stage patch selection module, $K$ selected patches in $B''$ are set as the nodes of the graph. Then, we employ the k-nearest neighbor algorithm ($KNN$, with $k = 8$) based on the spatial coordinates of the patch to model the connectivity of the node within a bag, obtaining the adjacency matrix $E = [E_{ij}] \in \mathbf{R}^{k \times k}$ of the graph, where $E_{ij}$ represents the edge between the $i^{th}$ patch and the $j^{th}$ patch. Formally, the adjacency matrix was written as:

$$E_{ij} = \begin{cases} 1, & \text{if } V_i \in KNN(V_j) \\ 0, & \text{otherwise} \end{cases}. \quad (3)$$

Based on the constructed patch-level graph, we analyze the graph feature through a simple GCN model:

$$V_g = \text{GCN}(V, E) = \{v_{g_i} \in \mathbf{R}^d, i = 1, 2 \dots K\}, \quad (4)$$

which aggregates the neighbor information of multiple slides, *i.e.*, providing their structural information.

*3.3.3 Hybrid Cross-Modal Feature Fusion.* Although $B''$ provides informative visual features, it lacks structural and global context. To integrate complementary cues, We propose a hybrid fusion module to integrate multimodal features. Visual features $B'' \in \mathbf{R}^{K \times d}$ are processed via multi-head self-attention, and the outputs from all heads are then normalized:

$$V_{att} = \text{LayerNorm}(\text{Softmax}(\frac{W_q B'' W_k B''^T}{\sqrt{d}})W_v B''), \quad (5)$$

where $W_q, W_k, W_v$ are learnable weights, LayerNorm($\cdot$) denotes the layer normalization operator, and Softmax($\cdot$) is the activation function. By employing this self-attention mechanism, the model effectively captures patch relationships, enhancing the correlation between local features and improving the model's ability to recognize fine-grained visual patterns. However, $V_{att}$ is lack of structural and global information due to its patch-level analysis strategy. To address this limitation, we aggregate the patch-level graph feature $V_g \in \mathbf{R}^{K \times d}$ through a cross-attention mechanism, which integrates global structural information with cross-modal feature interactions:

$$F_{vg} = \text{LayerNorm}(\text{CrossAtt}(V_{att}, V_g)), \quad (6)$$

where $F_{vg}$ is the fused feature, and $V_{att}$ serves as the query in the cross-attention mechanism. Then we concatenate $V_{att}$ and $F_{vg}$, utilizing learnable parameter $W$ to fuse these features, yielding the enhanced representation $F'_{vg} \in \mathbf{R}^{2K \times d}$. This process improves the model's understanding of complex spatial relationships, optimizes feature representations, and leads to improved classification accuracy and better generalization performance.

While the patch-level graph effectively compensates for structural information, global context remains limited since we only select $K$ patches from the WSIs. To address this limitation, we incorporate textual information to enhance global context representation. Given the visual features $F'_{vg} \in \mathbf{R}^{2K \times d}$ and text embeddings $T \in \mathbf{R}^{(C+t) \times d}$, we employ a cross-attention mechanism CrossAtt to fuse these multimodal features, with $T$ as the query. This integration generates a comprehensive representation $F_{vgt}$ that captures global contextual information and pathological knowledge.

$$F_{vgt} = \text{LayerNorm}(\text{CrossAtt}(T, F_{vg})). \quad (7)$$

### 3.4 Optimization

Given the dataset's class imbalance, we use FocalBCELoss [24], which adapts binary cross-entropy to focus on difficult examples in multi-label settings.

$$\mathcal{L} = \sum_{c=1}^{C} \alpha \cdot (1 - p_t^c)^\gamma \cdot \{- [y^c \log(p^c) + (1 - y^c) \log(1 - p^c)]\}, \quad (8)$$

where $C$ is the number of classes, $logists \in \mathbb{R}^C$ are the predicted logit vectors for all categories, and $y \in \{0, 1\}^C$ is the ground truth. The parameters are defined as: $\alpha \in [0, 1]$ is a weighting factor to balance positive and negative samples, and $\gamma \geq 0$ is the focusing parameter that reduces the loss contribution from easy examples.

## 4 Experiment

### 4.1 Dataset

To evaluate the effectiveness and generalizability of EmmPD, we conduct experiments on our in-house Placenta dataset and two public datasets: TCGA-Lung [46] and Camelyon+ [52]. Since no public placenta dataset exists, we construct a **Placenta** dataset covering three subtypes: avascular villi, placental infarction, and acute chorioamnionitis. Unlike TCGA-Lung, its labels are assigned at the patient level. Each patient contributes 2 to 4 WSIs scanned and may present with multiple disease types. In total, the dataset comprises 1,483 WSIs from 620 patients, with 284 diagnosed with avascular villi, 177 with placental infarction, and 375 with acute chorioamnionitis. Due to the limited number of patients for each disease, Placenta dataset was split at the patient level into training, validation, and test sets in a ratio of 7:1.5:1.5. **TCGA-lung** (TCGA) includes two disease subtypes: Lung Adenocarcinoma (LUAD, 492 slides) and Lung Squamous Cell Carcinoma (LUSC, 466 slides), with slide-level labels. The TCGA-Lung dataset is split into the training, validation, and test sets in 6:2:2 ratio. **Camelyon+** includes fours disease subtypes: negative (871 slides), micro (174), macro (251), ITC(54) diagnostic WSIs with slide-level labels. The Camelyon+ dataset was split 1:1:1 and evaluated with 3-fold cross-validation.

Table 1: Quantitative comparison with existing SOTA methods on three datasets. The best values are highlighted in bold.

| Method | Placenta | | | | TCGA-Lung | | | | Camelyon+ | | | | GPU | | Time |
| --- | --- | --- | --- | --- | --- | --- | --- | --- | --- | --- | --- | --- | --- | --- | --- |
| | Acc | F1 Score | ROC AUC | PR AUC | Acc | F1 Score | ROC AUC | PR AUC | Acc | F1 Score | ROC AUC | PR AUC | Train (GB) | Test (GB) | Time (ms) |
| ABMIL(2018) [14] | 0.6559 | 0.6160 | 0.5931 | 0.5485 | 0.8798 | 0.8797 | 0.9242 | 0.9092 | 0.8027 | 0.4894 | 0.7920 | 0.5387 | 1.75 | 0.95 | 0.24 |
| CLAMSB(2021) [30] | 0.6308 | 0.5617 | 0.6026 | 0.5693 | 0.8702 | 0.8698 | 0.9453 | 0.9376 | 0.6885 | 0.4984 | 0.7739 | 0.5064 | 1.57 | 1.09 | 1.65 |
| TransMIL(2021) [36] | 0.6380 | 0.5430 | 0.6132 | 0.5394 | 0.8750 | 0.8746 | 0.9468 | 0.9437 | 0.7807 | 0.5519 | 0.8031 | 0.5754 | 8.74 | 4.27 | 18.78 |
| DS-MIL(2021) [18] | 0.4588 | 0.6290 | 0.5825 | 0.5432 | 0.8077 | 0.8075 | 0.8874 | 0.8772 | 0.7800 | 0.4195 | 0.7508 | 0.5078 | 1.02 | 0.88 | 1.17 |
| DTFD-MIL)(2022) [55] | 0.4589 | 0.6324 | 0.5722 | 0.5602 | 0.8269 | 0.8267 | 0.9156 | 0.9056 | 0.7694 | 0.4229 | 0.7449 | 0.5092 | 1.56 | 1.01 | 0.07 |
| S4-Model(2023) [10] | 0.5986 | 0.3000 | 0.5046 | 0.4816 | 0.8750 | 0.8744 | 0.9402 | 0.9369 | 0.7983 | 0.4800 | 0.7419 | 0.5171 | 17.13 | 13.16 | 9.64 |
| WiKG(2024) [22] | 0.6380 | 0.5430 | 0.5677 | 0.5252 | 0.8462 | 0.8461 | 0.9089 | 0.8883 | 0.7892 | 0.5272 | 0.7920 | 0.5416 | 11.35 | 9.96 | 27.29 |
| PANTHER(2024) [44] | 0.6953 | 0.6104 | 0.6748 | 0.6610 | 0.9087 | 0.9085 | 0.9724 | 0.9741 | 0.7429 | 0.4300 | 0.7235 | 0.4731 | 1.76 | 1.37 | 0.18 |
| WSI-FIVE(2024) [20] | 0.6201 | 0.6214 | 0.5407 | 0.5215 | 0.8734 | 0.8708 | 0.9476 | 0.9375 | 0.7605 | 0.4990 | 0.7167 | 0.4986 | 11.17 | 3.02 | 404.68 |
| WSI-VQA(2024) [3] | 0.6380 | 0.5430 | - | - | 0.7644 | 0.7620 | - | - | 0.7288 | 0.4261 | - | - | 22.31 | 15.32 | 373.51 |
| SlideGCD(2024) [43] | 0.6129 | 0.6561 | 0.5794 | 0.5271 | 0.7788 | 0.7788 | 0.8507 | 0.8465 | 0.8008 | 0.4356 | 0.7361 | 0.4915 | 4.05 | 2.52 | 222.92 |
| MambaMIL(2024) [53] | 0.6129 | 0.6561 | 0.5988 | 0.5231 | 0.8654 | 0.8653 | 0.9353 | 0.9204 | 0.7930 | 0.5232 | 0.7936 | 0.5655 | 9.25 | 2.90 | 19.96 |
| PAM(2024) [12] | 0.6738 | 0.6431 | 0.6582 | 0.5796 | 0.8942 | 0.8939 | 0.9598 | 0.9615 | 0.8085 | 0.5152 | 0.8018 | 0.6096 | 2.16 | 1.82 | 27.6 |
| EmmPD | **0.7634** | **0.7295** | **0.7936** | **0.7215** | **0.9471** | **0.9471** | **0.9895** | **0.9900** | **0.8706** | **0.6750** | **0.8934** | **0.6752** | 2.58 | 1.68 | 19.71 |

Table 2: Comparison with existing SOTA methods using UNI as feature extractor. The best values are highlighted in bold.

| Modality | Method | Placenta | | | | TCGA-Lung | | | | Camelyon+ | | | |
| --- | --- | --- | --- | --- | --- | --- | --- | --- | --- | --- | --- | --- | --- |
| | | Acc | F1 Score | ROC AUC | PR AUC | Acc | F1 Score | ROC AUC | PR AUC | Acc | F1 Score | ROC AUC | PR AUC |
| image | ABMIL [14] | 0.6523 | 0.6756 | 0.7095 | 0.6172 | 0.9135 | 0.9133 | 0.9774 | 0.9784 | 0.8431 | 0.6030 | 0.8696 | 0.6580 |
| | CLAMSB [30] | 0.7419 | 0.7188 | 0.7797 | 0.7136 | 0.9230 | 0.9230 | 0.9845 | 0.9852 | 0.8166 | 0.5989 | 0.8701 | 0.6318 |
| | TransMIL [36] | 0.6703 | 0.6198 | 0.6977 | 0.6172 | 0.9135 | 0.9134 | 0.9713 | 0.9728 | 0.8292 | 0.5588 | 0.8668 | 0.6129 |
| | DS-MIL [18] | 0.4588 | 0.6290 | 0.6759 | 0.6377 | 0.9279 | 0.9278 | 0.9851 | 0.9857 | 0.8124 | 0.5352 | 0.8442 | 0.6080 |
| | DTFD-MIL [55] | 0.4588 | 0.6290 | 0.7500 | 0.6895 | 0.9231 | 0.9231 | 0.9874 | 0.9880 | 0.8287 | 0.5388 | 0.8289 | 0.6306 |
| | S4-Model [10] | 0.7025 | 0.5829 | 0.6760 | 0.6172 | 0.9375 | 0.9373 | 0.9838 | 0.9844 | 0.8654 | 0.6249 | 0.8530 | 0.6538 |
| | WiKG [22] | 0.7384 | 0.7045 | 0.7573 | 0.6924 | 0.9135 | 0.9128 | 0.9822 | 0.9828 | 0.8121 | 0.5751 | 0.8366 | 0.5978 |
| | PANTHER [44] | 0.6953 | 0.6104 | 0.6748 | 0.6610 | 0.9087 | 0.9085 | 0.9724 | 0.9741 | 0.7429 | 0.4300 | 0.7235 | 0.4731 |
| | WSI-FIVE [20] | 0.6667 | 0.5714 | 0.6840 | 0.6253 | 0.9375 | 0.9372 | 0.9814 | 0.9823 | 0.8496 | 0.5627 | 0.8557 | 0.6198 |
| | WSI-VQA [3] | 0.6846 | 0.6393 | - | - | 0.9271 | 0.9270 | - | - | 0.8367 | 0.6030 | - | - |
| | SlideGCD [43] | 0.6487 | 0.6316 | 0.6927 | 0.6254 | 0.9327 | 0.9325 | 0.9843 | 0.9850 | 0.8466 | 0.5862 | 0.8580 | 0.6137 |
| | MambaMIL [53] | 0.6989 | 0.6500 | 0.7360 | 0.6752 | 0.9375 | 0.9375 | 0.9886 | 0.9890 | 0.8488 | 0.6104 | 0.8722 | 0.6539 |
| | PAM [12] | 0.6201 | 0.6268 | 0.5477 | 0.4884 | 0.9423 | 0.9421 | 0.9829 | 0.9838 | 0.8199 | 0.5542 | 0.8634 | 0.6311 |
| image-text | MIZero [29] | 0.5771 | 0.5341 | 0.5333 | 0.5294 | 0.6154 | 0.6153 | 0.6606 | 0.6558 | 0.5127 | 0.3065 | 0.4786 | 0.2697 |
| | TOP [34] | 0.6295 | 0.4497 | 0.5373 | 0.5368 | 0.9130 | 0.9129 | 0.9609 | 0.9605 | 0.6314 | 0.1935 | 0.5129 | 0.2818 |
| | CONCH [28] | 0.5008 | 0.4821 | 0.5048 | 0.5259 | 0.6154 | 0.6151 | 0.6154 | 0.6010 | 0.6335 | 0.2305 | 0.5046 | 0.2597 |
| | EmmPD | **0.7634** | **0.7295** | **0.7936** | **0.7215** | **0.9471** | **0.9471** | **0.9895** | **0.9900** | **0.8706** | **0.6750** | **0.8934** | **0.6752** |

### 4.2 Implementation Details

We follow CLAM [30] for WSI preprocessing, including background removal, patching (512 × 512), and feature extraction. For TGCA-Lung cases with both 20× and 40× WSIs, only 40× images are used. We employ the pre-trained UNI encoder [5] to extract 1024-dimensional features per patch. To ensure a fair comparison, we not only reproduce all SOTA WSI classification approaches with their original feature extraction modules from the respective papers, but also evaluate them using UNI as feature extractor. Models are trained using Adam (batch size 1, learning rate $1 \times 10^{-4}$ with linear decay and early stopping). In patch selection, window size $w = 8 \times 8$, attention-based top $K = 4000$ patches are retained. The focal loss uses $\alpha = 0.25$, $\gamma = 2.0$. All experiments are conducted on one NVIDIA GeForce RTX 3090 GPU.

Table 3: Effectiveness evaluation of key modules of EmmPD on both datasets. The best values are highlighted in bold.

| Module | | | | Placenta | | | | TCGA-Lung | | | |
| --- | --- | --- | --- | --- | --- | --- | --- | --- | --- | --- | --- |
| 2DCom | AttSel | GCN | Text | Acc | F1 Score | ROCAUC | PRAUC | Acc | F1 Score | ROCAUC | PRAUC |
| ✓ | | | | 0.7168 | 0.6695 | 0.7469 | 0.6660 | 0.9231 | 0.9231 | 0.9777 | 0.9790 |
| | ✓ | | | 0.7061 | 0.6694 | 0.7218 | 0.6868 | 0.9183 | 0.9183 | 0.9690 | 0.9704 |
| ✓ | ✓ | | | 0.7240 | 0.7050 | 0.7580 | 0.6966 | 0.9327 | 0.9327 | 0.9827 | 0.9834 |
| ✓ | ✓ | ✓ | | 0.7348 | 0.7040 | 0.7705 | 0.6895 | 0.9375 | 0.9375 | 0.9832 | 0.9840 |
| ✓ | ✓ | | ✓ | 0.7419 | 0.7073 | 0.7757 | 0.7598 | 0.9375 | 0.9375 | 0.9819 | 0.9815 |
| ✓ | ✓ | ✓ | ✓ | **0.7634** | **0.7295** | **0.7936** | **0.7215** | **0.9471** | **0.9471** | **0.9895** | **0.9900** |

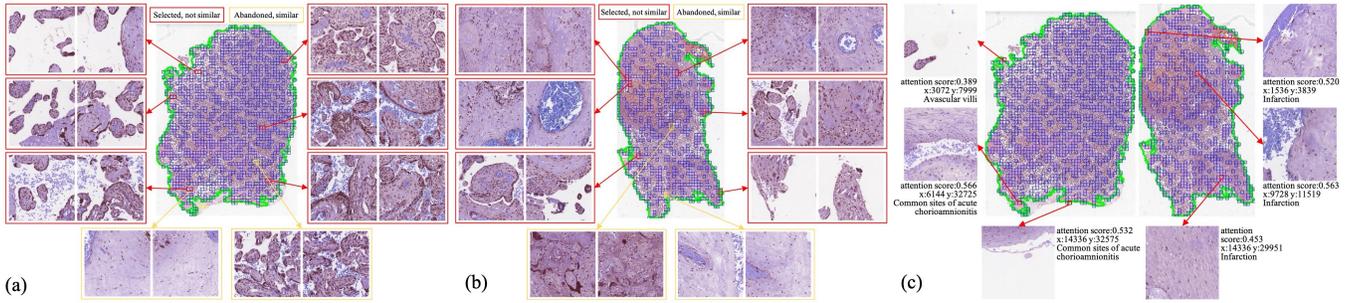

Figure 3: Retained patches from the two-stage patch selection module on two slides from a single patient with placental diseases. (a)(b) 2D visual token compression, (c) attention-based selection.

## 4.3 Metrics

To evaluate the performance of the proposed patient-level WSI classification model EmmPD, we report four metrics, namely Accuracy (Acc), F1 Score, Area Under the Receiver Operating Characteristic Curve (AUROC), Area Under the Precision-Recall Curve (PRAUC). The above indicators consider both patient- and category-level assessment, and are especially suitable for multi-label and unbalanced category distribution of pathological image classification tasks.

## 4.4 Evaluation Compared with SOTA Methods

To demonstrate the superiority of the proposed EmmPD in patient-level WSI classification, we conduct comprehensive experiments of EmmPD and state-of-the-art (SOTA) WSI classification methods. The experimental results in Table 1 are conducted with the same primary feature extractor as in their initial papers. We can clearly summarize from Table 1 that EmmPD significantly outperforms SOTA methods. EmmPD achieves an improvement of 5.29-30.46% in Acc, 5.32-42.95% in F1 Score, 2.97-28.90% in ROCAUC, 2.85-23.99% in PRAUC across three datasets. Most existing methods overlook joint modeling of local, structural, and global information, partly due to inefficient patch compression that hinders effective hybrid feature analysis. For fairness, we also compare the performance of our method with other methods with UNI image encoder as the primary feature extractor in Table 2. As we utilize image-text modalities as the model input, we further compare our approach with SOTA image-text based methods using UNI as feature extractor, *i.e.*, MIZero [29], TOP [34], and CONCH [28]. Their results also demonstrate that EmmPD achieves the top scores of all evaluation metrics across all datasets. We also benchmark GPU memory and inference time, and find that EmmPD achieves a strong trade-off between performance and efficiency, with significantly lower memory and inference time compared to transformer-based SOTA models. To validate robustness, we evaluate different LLMs and text encoders. Fig. 4(c) shows their overall impact is limited.

## 4.5 Ablation Study

*4.5.1 Impact of Key Modules.* To evaluate each component's contribution in EmmPD, we conduct ablation studies by progressively adding key modules: two-dimensional visual token compression (2DCom), attention-based patch selection (AttSel), GCN-based visual feature enhancement (GCN), and text-guided cross-attention (Text), as introduced in Fig. 2. Table 3 shows each module enhances overall performance. When only 2DCom, AttSel, or both are utilized for compression, we similarly select 4k patches, followed by a self attention layer and classification layer. Notably, the combination of both compression modules achieves the best results.

*4.5.2 Two-Stage Patch Selection.* From Table 3, we observe that the proposed Two-Stage Patch Selection (TSPS) module is both effective and efficient. Fig. 3 intuitively demonstrates that the first 2DCom module effectively discards average 60.67% redundant patches while retaining visually distinct ones. The subsequent attention-based selection can further filters out patches with low diagnostic relevance based on their score. We also conduct selection by existing three common methods, *i.e.*, random sampling 4k patches (random), selecting the top 9k patches based on their positional order (pos-9k), and one-dimensional visual token compression with 4k patches (1DCom), on both datasets. Table 4 outlines the results of different selection methods in EmmPD. While reducing patch count improves efficiency, these methods consistently underperform, highlighting their limited ability to retain diagnostic content. Patches in a WSI are highly correlated with neighbors, so window size $w$ directly affects patch selection quality. We conduct experiments with different $w$ in Table 5. Experiments show that overly small windows may discard subtle but critical patches, while overly large ones retain redundant, non-informative regions, both harming performance. As shown in Fig. 4(a,b), selecting an appropriate number of retained patches balances accuracy and efficiency. Retaining 4k patches achieves strong diagnostic performance with lower cost than 4.3k, while smaller $K$ values degrade accuracy due to insufficient information.

Table 4: Comparison of different sampling methods.

| Dataset | Method | Acc | F1 Score | ROCAUC | PRAUC |
|---|---|---|---|---|---|
| Placenta | random | 0.6703 | 0.6406 | 0.6854 | 0.5953 |
| | pos-9k | 0.6882 | 0.6360 | 0.7535 | 0.6460 |
| | 1DCom | 0.7348 | 0.6864 | 0.7620 | 0.6598 |
| | TSPS | **0.7634** | **0.7295** | **0.7936** | **0.7215** |
| TCGA-Lung | random | 0.9303 | 0.9305 | 0.9662 | 0.9576 |
| | pos-9k | 0.9231 | 0.9231 | 0.9772 | 0.9777 |
| | 1DCom | 0.9183 | 0.9183 | 0.9686 | 0.9639 |
| | TSPS | **0.9471** | **0.9471** | **0.9895** | **0.9900** |

Table 5: Comparison of different window size $w$ values.

| Dataset | $w$ | Acc | F1 Score | ROCAUC | PRAUC |
|---|---|---|---|---|---|
| Placenta | 6 | 0.7276 | 0.6960 | 0.7603 | 0.6558 |
| | 8 | **0.7634** | **0.7295** | **0.7936** | **0.7215** |
| | 10 | 0.7097 | 0.6747 | 0.7390 | 0.6623 |
| TCGA-Lung | 6 | 0.9399 | 0.9398 | 0.9846 | 0.9853 |
| | 8 | **0.9471** | **0.9471** | **0.9895** | **0.9900** |
| | 10 | 0.9375 | 0.9375 | 0.9709 | 0.9652 |

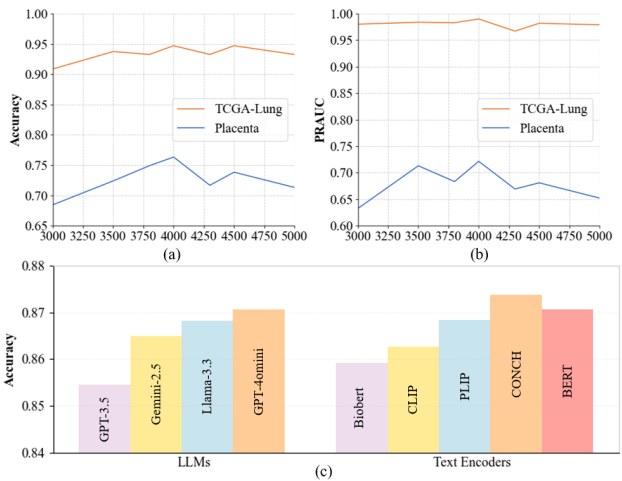

Figure 4: Ablation on the number of selected patches $K$ across all datasets w.r.t. (a) Accuracy and (b) PRAUC. (c) The impact of different LLMs and text encoders on model robustness.

Table 6: Comparison of GCN placement (before vs. after attention-based patch selection).

| Dataset | Method | Acc | F1 Score | ROCAUC | PRAUC |
|---|---|---|---|---|---|
| Placenta | before | 0.7240 | 0.6932 | 0.7746 | 0.7103 |
| | after | **0.7634** | **0.7295** | **0.7936** | **0.7215** |
| TCGA-Lung | before | 0.9183 | 0.9183 | 0.9772 | 0.9757 |
| | after | **0.9471** | **0.9471** | **0.9895** | **0.9900** |

Table 7: Comparison of different fusion methods.

| Dataset | Method | Acc | F1 Score | ROCAUC | PRAUC |
|---|---|---|---|---|---|
| Placenta | cat | 0.5735 | 0.5021 | 0.5024 | 0.4956 |
| | add | 0.5771 | 0.5280 | 0.4987 | 0.4837 |
| | ours | **0.7634** | **0.7295** | **0.7936** | **0.7215** |
| TCGA-Lung | cat | 0.2692 | 0.2692 | 0.1234 | 0.3275 |
| | add | 0.2861 | 0.2878 | 0.1111 | 0.3252 |
| | ours | **0.9471** | **0.9471** | **0.9895** | **0.9900** |

*4.5.3 Hybrid Multimodal Feature Fusion.* In this paper, we design a visual-GCN-text hybrid multimodal fusion strategy to enhance the model's representation ability. Table 1 has validated the effectiveness of both GCN- and text-based fusion modules. We further explore its effectiveness by locating it before or after attention-based patch selection module. The results in Table 6 prove that generating the graph after the attention-based patch selection module yields superior performance. This is because constructing the graph before this module includes excessive redundant nodes, which weakens the structural information. We also compare our fusion mechanism with two commonly used strategies, *i.e.*, simple addition (add) and concatenation (cat). They lead to a significant performance drop in Table 7, as such naive fusion strategies fail to fully exploit the complementary nature of visual, structural, and textual modalities.

More ablation studies and visualization results can be found in the supplementary materials.

## 5 Conclusion

In this study, we present **EmmPD**, an efficient multimodal framework for patient-level placental disease diagnosis using WSIs. EmmPD effectively addresses the dual challenges of computational burden and loss of global context in WSI classification by introducing a novel two-stage patch selection module and a hierarchical multimodal fusion strategy. By incorporating adaptive graph learning and textual medical knowledge, the framework enhances pathological feature representation and diagnostic accuracy. Extensive experiments on both a private dataset and two public datasets demonstrate the effectiveness and generalizability of EmmPD, setting a new SOTA in WSI-based disease diagnosis. And we will extend EmmPD to multi-center placental data analysis for broader downstream tasks.

## Acknowledgments

This work is supported by National Natural Science Foundation of China (Grant No. 62475072), Fundamental Research Funds for the Central Universities, the Science and Technology Commission of Shanghai Municipality (Grant No. 22S31905800, Grant No. 22DZ2229004).

# Supplementary of "Efficient Multi-Slide Visual-Language Feature Fusion for Placental Disease Classification"

Hang Guo*
East China Normal University
Shanghai, China
51285904046@stu.ecnu.edu.cn

Qing Zhang*
East China Normal University
Shanghai, China
qzhang@ce.ecnu.edu.cn

Zixuan Gao
East China Normal University
Shanghai, China
10232140401@stu.ecnu.edu.cn

Siyuan Yang
Nanyang Technological University
Singapore, Singapore
siyuan005@ntu.edu.sg

Shulin Peng
East China Normal University
Shanghai, China
10225304467@stu.ecnu.edu.cn

Xiang Tao
Obstetrics and Gynecology Hospital
of Fudan University
Shanghai, China
taoxiang1696@fckyy.org.cn

Ting Yu
Obstetrics and Gynecology Hospital
of Fudan University
Shanghai, China
yuting7410@fckyy.org.cn

Yan Wang
East China Normal University
Shanghai, China
ywang@cee.ecnu.edu.cn

Qingli Li†
East China Normal University
Shanghai, China
qlli@cs.ecnu.edu.cn

## A Discussion

### A.1 Visualization of features from Modules

We visualize the features extracted from different modules in the proposed EMMPD in Fig. 1, including attention score maps from attention based patch selection, visual feature from self attention, graph feature from GCN-based visual feature enhancement, and visual-graph feature from GCN-guided cross attention. It can be observed that, visual feature focuses on detail information while graph feature captures structural information. Combining both features enhances key patches relevant to disease diagnosis while preserving the detail information. In the enhanced feature map, the key patches do not exhibit a strong regional distribution. This is primarily because a large number of redundant patches have already been filtered out based on the attention score map. Therefore, patches that receive lower attention scores, referred to as red patches, are no longer present in the final feature map, as they have been discarded during selection process.

### A.2 Visualization of Selected Patches in TCGA-Lung

Fig. 2 presents the patches selected by the proposed two-stage patch selection module on the TCGA-Lung dataset. A large number of redundant patches are discarded through two-dimensional visual

*Both authors contributed equally to this research.
†Corresponding author.

Table 1: comparison of different sample number $K$ on both datasets. The best values are highlighted in bold.

| Dataset | $K$ | Acc | F1 Score | ROCAUC | PRAUC |
|---|---|---|---|---|---|
| Placenta | 3k | 0.6846 | 0.6641 | 0.7312 | 0.6326 |
| | 3.5k | 0.7240 | 0.6980 | 0.7845 | 0.7131 |
| | 3.8k | 0.7491 | 0.7154 | 0.7423 | 0.6833 |
| | 4k | **0.7634** | **0.7295** | **0.7936** | **0.7215** |
| | 4.3k | 0.7168 | 0.6973 | 0.7513 | 0.6693 |
| | 4.5k | 0.7384 | 0.7045 | 0.7401 | 0.6810 |
| | 5k | 0.7133 | 0.68 | 0.7378 | 0.6522 |
| TCGA-Lung | 3k | 0.9087 | 0.9091 | 0.9791 | 0.9802 |
| | 3.5k | 0.9375 | 0.9375 | 0.9830 | 0.9839 |
| | 3.8k | 0.9327 | 0.9327 | 0.9826 | 0.9829 |
| | 4k | **0.9471** | **0.9471** | **0.9895** | **0.9900** |
| | 4.3k | 0.9327 | 0.9327 | 0.9757 | 0.9671 |
| | 4.5k | **0.9471** | **0.9471** | 0.9828 | 0.9819 |
| | 5k | 0.9327 | 0.9327 | 0.9809 | 0.9791 |

Table 2: comparison of different parameter $\alpha$ of the loss function on both datasets. The best values are highlighted in bold.

| Dataset | $\alpha$ | Acc | F1 Score | ROCAUC | PRAUC |
|---|---|---|---|---|---|
| Placenta | 0.05 | 0.7312 | 0.7059 | 0.7558 | 0.6806 |
| | 0.15 | 0.7168 | 0.6877 | 0.7371 | 0.6808 |
| | 0.25 | **0.7634** | **0.7295** | **0.7936** | **0.7215** |
| | 0.35 | 0.6846 | 0.6364 | 0.7145 | 0.6449 |
| | 0.45 | 0.6882 | 0.6506 | 0.7100 | 0.6435 |
| TCGA-Lung | 0.05 | 0.9062 | 0.9060 | 0.9782 | 0.9785 |
| | 0.15 | 0.9135 | 0.9135 | 0.9711 | 0.9721 |
| | 0.25 | **0.9471** | **0.9471** | **0.9895** | **0.9900** |
| | 0.35 | 0.9327 | 0.9320 | 0.9795 | 0.9750 |
| | 0.45 | 0.9231 | 0.9231 | 0.9784 | 0.9774 |

Table 3: comparison of different parameter $\gamma$ of the loss function on both datasets. The best values are highlighted in bold.

| Dataset | $\gamma$ | Acc | F1 Score | ROCAUC | PRAUC |
|---|---|---|---|---|---|
| Placenta | 1.0 | 0.6882 | 0.6790 | 0.7297 | 0.6426 |
| | 2.0 | **0.7634** | **0.7295** | **0.7936** | **0.7215** |
| | 3.0 | 0.7204 | 0.6803 | 0.7389 | 0.6459 |
| | 4.0 | 0.7348 | 0.7016 | 0.7983 | 0.7365 |
| | 5.0 | 0.6918 | 0.6228 | 0.7642 | 0.6852 |
| TCGA-Lung | 1.0 | 0.9279 | 0.9279 | 0.9704 | 0.9683 |
| | 2.0 | **0.9471** | **0.9471** | **0.9895** | **0.9900** |
| | 3.0 | 0.9231 | 0.9231 | 0.9742 | 0.9691 |
| | 4.0 | 0.9279 | 0.9279 | 0.9784 | 0.9772 |
| | 5.0 | 0.9375 | 0.9375 | 0.9827 | 0.9828 |

Table 4: Comparison of different LLMs and text encoders

| Dataset | Method | Acc | F1 Score | ROC AUC | PR AUC |
|---|---|---|---|---|---|
| LLM | GPT-3.5 [1] | 0.8546 | 0.6345 | 0.8772 | 0.6503 |
| | Gemini-2.5 [8] | 0.8649 | 0.6398 | 0.8715 | 0.6572 |
| | Llama-3.3B [9] | 0.8683 | 0.6513 | 0.8681 | 0.6599 |
| | GPT-4omini(ours) [6] | 0.8706 | 0.6750 | 0.8934 | 0.6752 |
| Text-Encoder | Biobert [4] | 0.8592 | 0.6078 | 0.8771 | 0.6463 |
| | CLIP [7] | 0.8626 | 0.6319 | 0.8739 | 0.6655 |
| | PLIP [3] | 0.8684 | 0.6521 | 0.8932 | 0.6881 |
| | CONCH [5] | 0.8737 | 0.6425 | 0.8779 | 0.6747 |
| | BERT(ours) [2] | 0.8706 | 0.6750 | 0.8934 | 0.6752 |

### B.3 Robustness of the Model

We conduct ablation studies on both LLMs and text encoders in Table 4. Results show the overall impact of different LLMs and text encoders is not significant.

token compression and attention-based selection, enhancing the model's efficiency.

## B Statistical Analysis

### B.1 Number of Selected Patches from WSIs

We report the quantitative evaluation results for different values of the selected patch number $K$ across all evaluation metrics in Table 1, demonstrating that our method achieves optimal performance when $K = 4000$.

### B.2 Hyperparameter in Loss Function

We adjusted the FocalBCELoss parameters $\alpha$ and $\gamma$ to validate their impact, in Table 2 and Table 3. Extensive experiments prove that $\alpha = 0.25$ optimally balances the class weights between positive and negative samples, and $\gamma = 2.0$ most effectively reduced the loss contribution from well-classified easy samples.

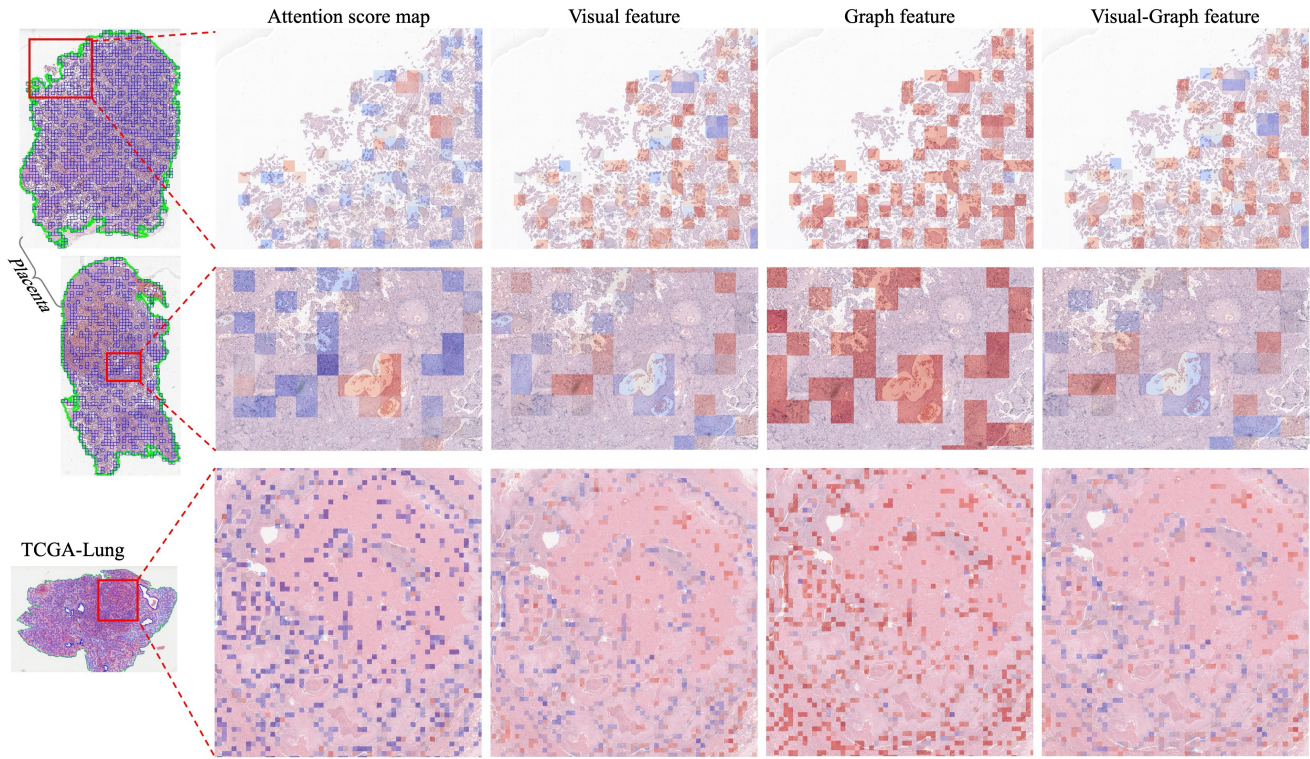

Figure 1: Visualization of features from different modules in EmmPD.

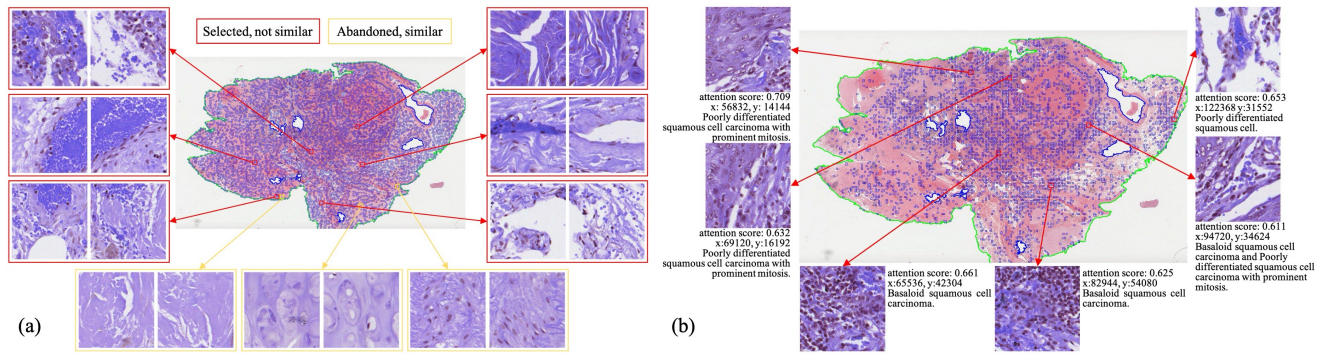

Figure 2: Retained patches from the two-stage patch selection module on TCGA-Lung dataset. (a) 2D visual token compression, (b) attention-based selection.